\documentclass{article} 
\usepackage{iclr2027_conference,times}

\usepackage{amsmath,amsfonts,bm}

\def\eqref#1{equation~\ref{#1}}

\def\1{\bm{1}}

\DeclareMathAlphabet{\mathsfit}{\encodingdefault}{\sfdefault}{m}{sl}
\SetMathAlphabet{\mathsfit}{bold}{\encodingdefault}{\sfdefault}{bx}{n}

\usepackage{graphicx}
\usepackage{subcaption}
\usepackage{bbm}
\usepackage{bm}
\usepackage{multirow}
\usepackage{amsmath}
\usepackage{amsthm}
\usepackage{hyperref}
\usepackage{wrapfig}
\usepackage{placeins}
\usepackage{amssymb}
\usepackage[capitalize,noabbrev]{cleveref}
\usepackage{pifont}

\newcommand{\cmark}{\ding{51}}
\newcommand{\xmark}{\ding{55}}
\theoremstyle{plain}
\newtheorem{theorem}{Theorem}[section]
\newtheorem{proposition}[theorem]{Proposition}

\theoremstyle{definition}

\theoremstyle{remark}

\usepackage[textsize=tiny]{todonotes}

\usepackage[utf8]{inputenc} 
\usepackage[T1]{fontenc}    
\usepackage{hyperref}       
\usepackage{url}            
\usepackage{booktabs}       
\usepackage{amsfonts}       
\usepackage{nicefrac}       
\usepackage{microtype}      
\usepackage{xcolor}         

\title{Causal Front-Door Adjustment for Robust Jailbreak Attacks on LLMs}

\author{
Yao Zhou$^{1,2}$\thanks{Equal contribution.}
\quad
Zeen Song$^{1,2}$\textsuperscript{*}
\quad
Wenwen Qiang$^{1,2}$\thanks{Corresponding author: \texttt{qiangwenwen@iscas.ac.cn}}
\quad
Fengge Wu$^{1,2}$
\quad
Changwen Zheng$^{1,2}$
\\[4pt]
$^{1}$National Key Laboratory of Space Integrated Information System,\\[-1pt]
\phantom{$^{1}$}Institute of Software, Chinese Academy of Sciences, Beijing, China\\[2pt]
$^{2}$University of Chinese Academy of Sciences, Beijing, China
}

\iclrfinalcopy

\begin{document}

\maketitle
\lhead{}

\begin{abstract}
Jailbreak attacks on safety-aligned LLMs are increasingly effective, yet attack success alone does not reveal whether compliance is driven by preserved task intent, suppressed safety signals, or attack-specific artifacts. We use Pearl’s front-door adjustment to motivate separating task-relevant semantics from safety-related variation. We propose \textbf{Causal Front-Door Adjustment Attack} (\textbf{CFA}$^2$), a white-box framework that operationalizes this perspective in LLM latent space using Sparse Autoencoders, contrastive datasets, and weight orthogonalization. After offline calibration, CFA$^2$ enables efficient generation without per-prompt iterative optimization, performs strongly on HarmBench across four open-source LLMs, and provides source-side signals for same-family black-box prompt transfer.
\end{abstract}

\section{Introduction}
Large Language Models (LLMs) \cite{qi2024safety,chao_jailbreaking_2024,liu_autodan_2024} have revolutionized human-computer interaction, yet they expose significant safety vulnerabilities. Even models aligned with rigorous safety protocols can be manipulated to generate harmful content through jailbreaking attacks~\cite{zhang2025boosting,liu_flipattack_2024,li2023deepinception}.

Recent jailbreak methods have demonstrated strong effectiveness through diverse intervention interfaces, including adversarial suffix optimization, semantic prompt rewriting, output-space manipulation, and representation-level steering~\cite{zou2023universaltransferableadversarialattacks,chao_jailbreaking_2024,liu_autodan_2024}. However, high attack success does not by itself explain why a model complies. Since many methods operate by adding adversarial suffixes, role-playing contexts, output prefixes, or other auxiliary signals, the resulting input differs from the original harmful request in more than one aspect. Thus, attack success alone cannot reveal whether compliance is driven by preserved task intent, suppressed safety signals, or other reasons.

Motivated by this challenge, we investigate the underlying generation mechanism of LLMs through a causal lens. Our approach leverages the insight that safety alignment may act as a superficial constraint rather than erasing underlying knowledge~\cite{zhou2023lima,wei2023jailbroken}. Specifically, pre-trained models possess the capability to answer harmful queries, but this capability is suppressed by a latent safety mechanism introduced during alignment. In causal terms, jailbreak failures occur not because the model ``cannot'' answer, but because the safety mechanism acts as a confounder that interferes with the output.

Building on this insight, we formulate a Structural Causal Model (SCM), as illustrated in Figure~\ref{fig:SCM}. Let $X$ denote the harmful query and $Y$ the response. We introduce an unobserved node $U$ to represent the internal safety mechanism. Crucially, we model $U$ as a latent confounder: it influences how the query is processed (path $U \to A$, $A$ is the model's internal latent representation of $X$) and directly biases the output towards refusal ($U \to Y$). This confounding effect explains why simple optimization fails; that is, it cannot distinguish confounded associations between the model's true capability and the safety interference.

To account for this confounding structure, we leverage Pearl's Front-Door Adjustment~\cite{pearl2009causality}. We introduce an observable mediator $S$, representing the core task semantics of the query. We instantiate $S$ as the task-relevant semantic component of the query, which is designed to mediate the query-response pathway while minimizing its dependence on safety-related variation. Under this construction, front-door adjustment provides a principled way to estimate the interventional effect of the task pathway despite the unobserved safety mechanism $U$. Finally, we propose the \textbf{C}ausal \textbf{F}ront-Door \textbf{A}djustment \textbf{A}ttack ({\textbf{CFA}$^2$}), a framework that translates this causal theory into a practical, efficient attack algorithm. Specifically, CFA$^2$ implements this framework via a two-stage pipeline: First, employing Sparse Autoencoders with contrastive analysis to extract interpretable features and separate approximately invariant task-related features from defense mechanisms; and second, applying weight orthogonalization to suppress the identified defense-related direction, yielding a fixed parameter-level intervention with $O(1)$ attack-specific overhead with respect to per-prompt optimization rounds, followed by standard autoregressive decoding.

Our key contributions: (1) We propose a \textbf{causal formulation of jailbreak intervention} that models safety-related mechanisms as latent factors entangled with task-relevant semantics. Rather than treating jailbreak success purely as an input-output phenomenon, our formulation characterizes the transition from refusal to compliance as a problem of separating the task pathway from safety-induced interference. 
(2) We introduce \textbf{CFA$^2$}, an offline-calibrated and inference-efficient jailbreak algorithm. CFA$^2$ uses SAEs to identify approximately invariant task-related features. Crucially, we employ \textit{Weight Orthogonalization} to project out estimated defense-related directions. This yields $O(1)$ attack-specific overhead with respect to the number of per-prompt optimization steps.
(3) We provide \textbf{extensive empirical evidence} across effectiveness, efficiency, and transferability. CFA$^2$ performs strongly on HarmBench across four open-source LLMs, outperforming prompt-level attacks as well as stronger white-box or internal-state baselines such as ORTHO, DSN, and SM. CFA$^2$ further enables standard autoregressive generation without per-prompt iterative optimization and provides useful source-side signals for same-family black-box prompt transfer.


\section{Related Work}
\textbf{LLM Jailbreak.}
Existing jailbreak methodologies can be broadly grouped into three categories: optimization-based adversarial attacks~\cite{zou2023universaltransferableadversarialattacks,liu_autodan_2024,zhang2025boosting,zhou2025don,jin2024jailbreaking}, semantic manipulation~\cite{chao_jailbreaking_2024,li2023deepinception,mehrotra2024tree,chen_pandora_2024,li_drattack_2024,shah_scalable_2023,yong_low-resource_2024,deng_multilingual_2024,yuan_gpt-4_2024,yao_fuzzllm_2024}, and fine-tuning or decoding interventions~\cite{zhang_jailbreak_2024,huang_catastrophic_2023}.
These methods manipulate prompts, decoding procedures, or model internals to increase harmful compliance. Representation-level attacks are most relevant to our setting. Refusal Direction Orthogonalization (ORTHO) identifies a dense refusal-related direction~\cite{arditi2024refusal}. Don't Say No (DSN) suppresses refusal behavior through optimization-based intervention~\cite{zhou2025don}, while Internal Stance Manipulation (SM) modifies internal states to shift the model toward compliance~\cite{fu2025jailbreak}. Although effective, these methods typically identify refusal or behavior related directions without explicitly controlling for task-semantic variation. More broadly, many jailbreak methods alter multiple aspects of the original request or generation process. This limits their interpretability as red-teaming tools, since the observed compliance does not clearly indicate whether success arises from preserved task intent, suppressed refusal features, or attack-specific artifacts.

\textbf{Representation Learning and Mechanistic Interpretability.}
Representation learning provides a useful framework for analyzing and manipulating internal LLM states.
Representation Engineering (RepE), for example, steers model behavior through dense activation directions~\cite{zou2023representation}, dense representations can be limited by polysemanticity and feature superposition, where dense representations entangle multiple semantic factors~\cite{elhage2022toy}.
To address this, SAE utilizes high-dimensional sparse projections to resolve feature superposition. Recent work has used SAEs to identify localized features associated with semantic concepts and model behaviors~\cite{huben2023sparse,bricken2023towards,templeton2024scaling}.
However, current SAE research is predominantly limited to descriptive analysis or benign steering (e.g., enhancing honesty)~\cite{subramani2022extracting}, leaving their potential for adversarial red-teaming comparatively unexplored.

\section{Problem Formulation and Causal Analysis}
\label{sec:Problem Formulation and Analysis}


In this section, we establish the theoretical foundation for our proposed method. We begin by defining the objective of jailbreak attacks and introducing the mathematical notations in Section~\ref{sec:problem_def}. Building on this formulation, we then transition to a causal perspective in Section~\ref{sec:causal_analysis}. Here, we construct a Structural Causal Model (SCM) to characterize the interference of unobserved safety mechanisms and apply the Front-Door Adjustment to theoretically derive a robust jailbreaking strategy.

\subsection{Problem Formulation}
\label{sec:problem_def}

In this subsection, we formally define the objective of the LLM jailbreak attack. Modern LLMs operate under an inherent tension: they retain vast knowledge and instruction-following capabilities from pre-training, but are constrained by safety alignment. A \textbf{jailbreak attack} is an adversarial attempt to bypass these constraints, aiming to elicit harmful, unethical, or restricted responses.

Let $\mathcal{D}=\{x_i\}_{i=1}^N$ be a dataset of harmful requests drawn from a distribution $P(x)$. Given a pre-trained LLM $\pi_\theta$, the model defines a conditional distribution $\pi_\theta(y \mid x)$ over the response space $\mathcal{Y}$. For any input $x$, the response $y$ is sampled autoregressively, i.e., $y \sim \pi_\theta(\cdot \mid x)$. We further define a binary judge function $g: \mathcal{Y} \to \{0,1\}$, which determines whether a response $y$ is a successful jailbreak ($g(y)=1$) or a harmless refusal ($g(y)=0$). 

The objective of a jailbreak attack is to maximize the expected success rate, formally defined as:
\begin{equation}
    P_{\text{success}} \triangleq \mathbb{E}_{x\sim P(x)}\mathbb{E}_{y\sim \pi_\theta (\cdot|x)}\big[ \mathbbm{1}[g(y)=1] \big],
\end{equation}
where $\mathbbm{1}[\cdot]$ denotes the indicator function. Intuitively, this objective aims to find the optimal strategy (e.g., input perturbations) to maximize the likelihood of harmful outputs. This essentially attempts to shift the model's output distribution from the safe region toward the harmful space.

\begin{wrapfigure}{r}{0.42\textwidth}
    \vspace{-12pt}
    \centering
    \includegraphics[width=0.38\textwidth]{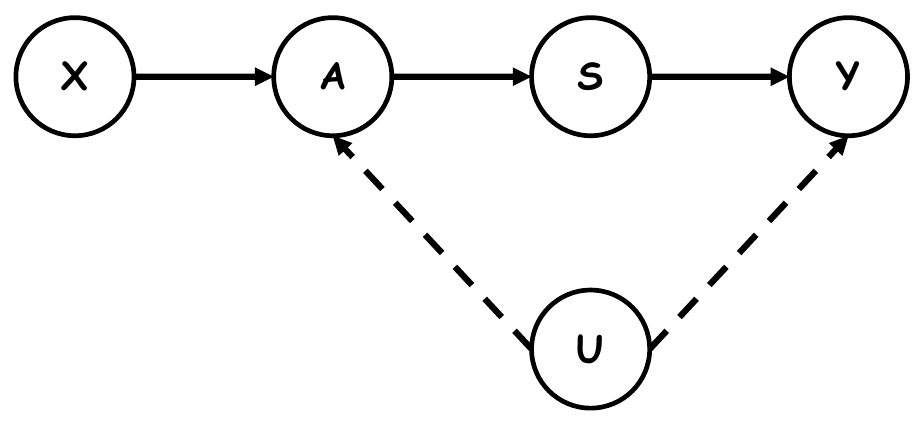}
    \caption{\textbf{Jailbreaking SCM.} $X$: query, $A$: representation, $S$: semantics, $Y$: harmful response, and $U$: latent safety mechanism.}
    \label{fig:SCM}
    \vspace{-12pt}
\end{wrapfigure}

\subsection{Causal Analysis}
\label{sec:causal_analysis}
From the problem formulation, our objective is to jailbreak a safety-aligned language model. In principle, a pre-trained model possesses the inherent capability to generate practical responses to arbitrary queries, regardless of their safety status. However, safety alignment introduces a constraint where the model produces refusals or templated responses for specific inputs. Crucially, such refusals do not imply a lack of underlying capability; rather, they indicate the presence of unobserved internal mechanisms (i.e., safety guardrails) that actively suppress the answer. To understand this suppression and recover the model's answering capability, we model the interactions between queries, responses, and safety mechanisms through a causal lens.

\textbf{The Structural Causal Model (SCM).}
We denote the input query by $X$, the response by $Y$, and the unobserved internal safety mechanism by $U$. Additionally, we introduce $A$ to represent the model's internal latent representation of $X$.


In our causal graph (see Figure~\ref{fig:SCM}), the path $X \to A \to Y$ represents the model's inherent task capability at a coarse level, e.g., the ideal mapping from query to representation to response in the absence of constraints. However, safety alignment introduces a confounder $U$. The safety mechanism $U$ influences the internal representation $A$ (making observed representations a mixture of task and safety signals, $U \to A$) and explicitly biases the output $Y$ towards refusal ($U \to Y$)~\cite{arditi2024refusal,wolf2024fundamentallimitationsalignmentlarge}.
Consequently, the observed conditional distribution $P(Y \mid X)$ is a mixture of the model's potential capability and the safety mechanism's intervention. To isolate the model's original answering capability, we introduce a conceptual mediator variable $S$, representing the core task semantics within $X$. We posit that the causal effect of task information on $Y$ is strictly mediated through $S$, and we assume that \(S\) intercepts the directed effect of \(A\) on \(Y\). Structurally, this decomposes the generation process into two stages: extracting the core task representation ($X \to A \to S$) and generating the response conditioned on this task-specific semantic ($S \to Y$). By introducing $S$, we isolate the mapping from $X$ to $Y$ driven solely by task information, effectively filtering out the non-task safety activations present in $A$.

\textbf{Front-door Adjustment for Confounding Resolution.}
To formally characterize the genuine influence of the input on the response while shielding the process from the safety mechanism $U$, we employ the causal intervention operator $do(\cdot)$. The objective of jailbreak is defined as the interventional distribution $P(Y \mid do(A))$. It represents the response distribution under an intervention that externally sets \(A\), thereby removing the confounding influence on \(A\).

Pearl's Front-door Criterion~\cite{pearl2009causality} provides the theoretical foundation for identifying $P(Y \mid do(A))$ even when the confounder $U$ is unobserved. Specifically, because the mediator $S$ shields the direct influence of input features on $Y$, and the path from $A$ to $S$ is not intercepted by $U$, the causal intervention probability can be derived as:
\begin{equation}
\label{eq:causal_optimization}
    P(Y=\bm{y}|do(A=\bm{a}))={\textstyle \sum_{s}}P(S=s|A=\bm{a})\\
{\textstyle \sum_{\bm{a}^\prime}}P(Y=\bm{y}|A=\bm{a}^\prime,S=s)P(A=\bm{a}^\prime)
\end{equation}


Intuitively, the outer summation captures the activation of the task semantics $S$ by the input $\bm{a}$. The inner expectation marginalizes over the input space $\bm{a}'$, thereby adjusting for the spurious association induced by the unobserved confounder $U$. This implies that we do not need to model the defense mechanism $U$ explicitly; instead, we can use the mediator \(S\) to identify the task-mediated pathway through which the effect of \(do(A)\) can be recovered.


\textbf{Inspiration and Challenges.}
The front-door adjustment framework provides a causal view for separating task semantics from
safety-related confounding. However, applying it to LLM jailbreak intervention requires addressing
three challenges. First, \textbf{Representation Identification}: the mediator $S$ should be located in
high-dimensional, polysemantic activations. Second, \textbf{Estimation Efficiency}: the required
marginalization over background representations $A'$ is infeasible to compute directly. Third,
\textbf{Operationalization}: the causal adjustment should be translated into a lightweight model-level
intervention that preserves task semantics while suppressing safety-related variation.

\section{Methodology}
\begin{figure*}[t]
    \centering
    \includegraphics[width=\linewidth]{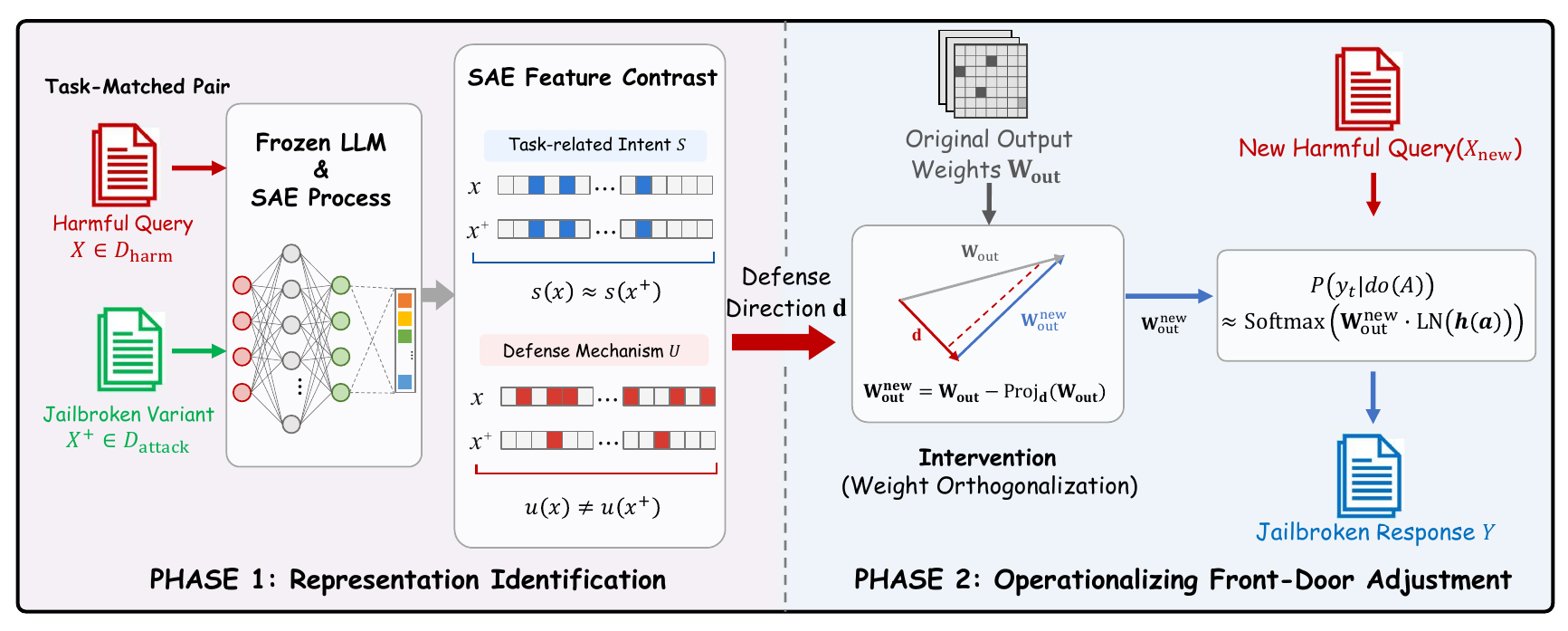}
    \caption{
    Overview of the Causal Front-Door Adjustment Attack (CFA$^2$) framework. CFA$^2$ uses task-matched harmful/jailbroken pairs to identify approximately invariant task-related features and a defense-related direction $\mathbf{d}$. It then projects $\mathbf{W}_{\mathrm{out}}$ onto the orthogonal complement of $\mathbf{d}$, yielding modified weights $\mathbf{W}_{\mathrm{out}}^{\mathrm{new}}$. This parameter-level intervention suppresses the identified defense-related contribution while retaining task-related information, enabling standard autoregressive generation without per-prompt iterative optimization.}
    \label{fig:method_arch}
    \vspace{-5pt}
\end{figure*}

To address the aforementioned challenges, we propose a novel method termed \textbf{C}ausal \textbf{F}ront-Door \textbf{A}djustment \textbf{A}ttack ({\textbf{CFA}$^2$}). The central idea of CFA$^2$ is to leverage causal inference principles to design a causality-inspired jailbreak strategy for LLM.
The method comprises two stages. In the first stage, we employ a Sparse Autoencoder (SAE) to decompose the LLM hidden representations into sparse latent features, and use task-matched contrastive analysis to identify approximately invariant task-related features and defense-related variation. In the second stage, we translate the identified defense-related variation into a residual-stream direction and apply weight orthogonalization to suppress its contribution to the output mapping. The resulting parameter-level intervention provides a tractable approximation to the marginalization effect motivated by the front-door formulation and is then used with standard autoregressive decoding.
\subsection{Identification of the Front-Door Mediator}
A central challenge in implementing front-door adjustment is the accurate identification and formalization of the mediator $S$ from observed data. In LLM tasks, the model's internal reasoning logic is highly compressed and encoded within dense, high-dimensional hidden activations~\cite{geva2021transformer,park2023linear,meng2022locating}. The inherent polysemanticity and superposition of these representations make it extremely challenging to extract the core task representation.
To address this challenge, we draw inspiration from recent advances in SAE \cite{ng2011sparse,ferrando2024know}, which have been shown to disentangle complex hidden activations into interpretable and localized sparse features.     
Specifically, an SAE is constructed as a neural network with an encoder function $\mathbf{z} = \phi(\mathbf{W}\mathbf{x} + \mathbf{b})$ and a decoder function $\hat{\mathbf{x}} = \mathbf{W}' \mathbf{z} + \mathbf{b}'$where \(\mathbf{x} \in \mathbb{R}^d\) denotes the dense hidden activation vector of the LLM, \(\mathbf{z} \in \mathbb{R}^k\) is the sparse latent representation, while \(\mathbf{b} \in \mathbb{R}^k\) and \(\mathbf{b}' \in \mathbb{R}^d\) correspond to the bias vectors, and \(\phi\) are nonlinear activation functions. 
The training objective combines a reconstruction loss with a sparsity-inducing penalty, where $\lambda$ controls the trade-off between accuracy and sparsity. The final optimization target is formulated as:
\begin{equation}
    \min_{\mathbf{W}, \mathbf{W}', \mathbf{b}, \mathbf{b}'} \mathcal{L} = \|\mathbf{x} - \hat{\mathbf{x}}\|_2^2 + \lambda \|\mathbf{z}\|_1
    \label{eq:sae_loss}
\end{equation}
Through this formulation, the autoencoder learns to reconstruct hidden activations while forcing most latent dimensions in \(\mathbf{z}\) to remain inactive via L1 regularization. By applying an SAE to the hidden layers of an LLM, the dense activation latent vectors are projected into a sparse latent space, where each dimension corresponds to a candidate feature with clearer semantic interpretation. This transformation provides a principled means of searching for front-door variables: candidate variables are defined as the sparse latent dimensions extracted by the SAE, which can then be systematically examined to test whether they mediate the causal pathway between input prompts and model outputs. In this way, SAE serves not only as a dimensionality reduction tool but also as a bridge between raw model activations and causally meaningful variables.

To establish a theoretical basis for identifying the desired latent factors from SAE representations, we invoke the identifiability theory from the framework of nonlinear Independent Component Analysis (ICA) and disentangled representation learning \cite{khemakhem2020variational,von2021self}.

\begin{figure*}[t]
    \centering
    \includegraphics[width=\linewidth]{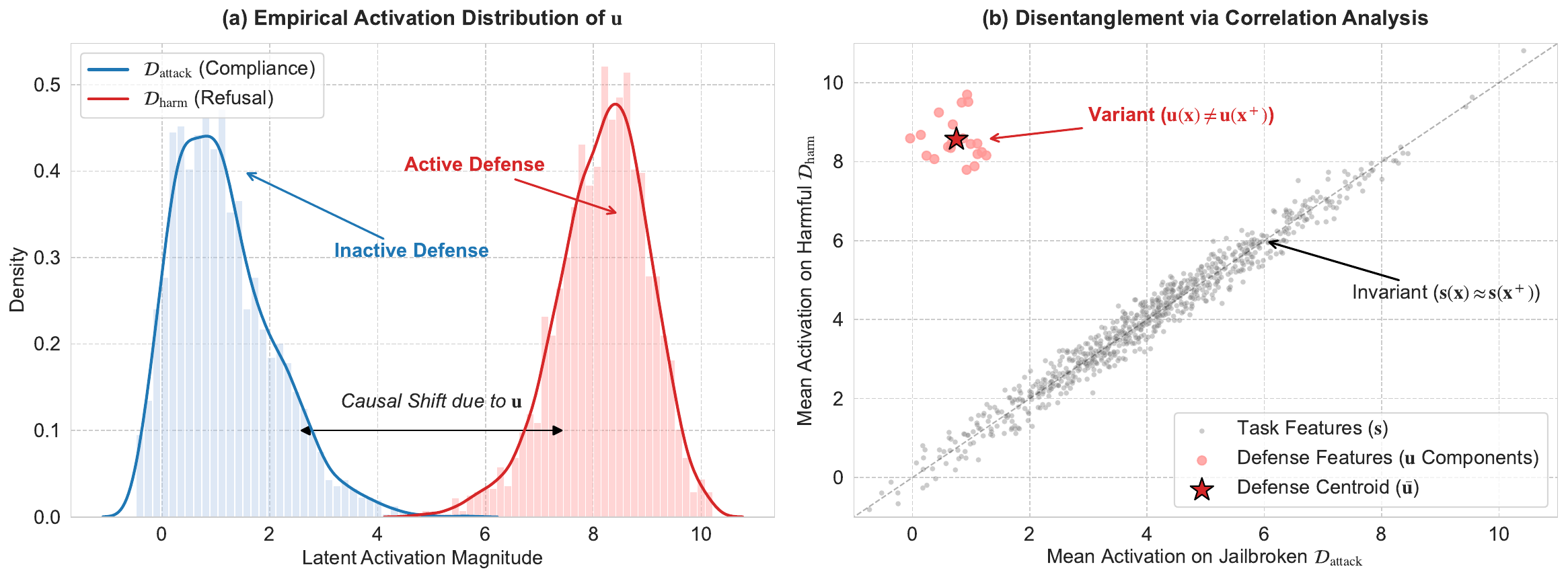}
    \caption{Validation of Disentanglement (Proposition 4.2).
(a) Evidence of Style Variance: The activation distribution of $\bm{u}$ shows a distinct causal shift between the refusal state ($\mathcal{D}_{harm}$) and compliance state ($\mathcal{D}_{attack}$), confirming $\bm{u}(\bm{x}) \neq \bm{u}(\bm{x}^+)$.
(b) Evidence of Content Invariance: While the defense mechanism (red star) acts as an outlier, the core task semantics (gray points, $\mathbf{s}$) remain aligned along the invariance line ($y=x$).}
    \label{fig:theorem2_evidence}
    \label{fig:variance}
\vspace{-5pt}
\end{figure*}

\begin{proposition}[Identifiability of Latent Causal Factors]
\label{prop:1}
        Let the data generating process be $\bm{x} = \bm{g}(\bm{z})$, where $\bm{g}: \mathcal{Z} \to \mathcal{X}$ is a smooth diffeomorphism and $\bm{z} \in \mathbb{R}^{d_z}$ are latent factors. Assume the prior $p(\bm{z}|y)$ conditioned on the response $y$ follows a conditionally factorial exponential family with sufficient statistics $\bm{T}(\bm{z})$ and parameters $\bm{\lambda}(y)$.
    If the learned representation $\bm{a} = \bm{f}(\bm{x})$ satisfies the following conditions:
    
    (1) \textbf{Reconstruction Consistency}: The SAE reconstructs the input density, i.e., $p_{\text{SAE}}(\bm{x}) = p_{\text{true}}(\bm{x})$.
    
    (2) \textbf{Task Sufficiency}: The SAE features preserve task information:
$D_{KL}(p_{\text{SAE}}(y|\bm{x}) \| p_{\text{true}}(y|\bm{x})) = 0$.
    
    (3) \textbf{Variability}: The responses are diverse enough that
$\bm{L} = [\bm{\lambda}(y_1) - \bm{\lambda}(y_0), \dots, \bm{\lambda}(y_k) - \bm{\lambda}(y_0)]$
has rank $k$, the dimension of the sufficient statistics.
    
    (4) \textbf{Sparsity Prior}: The true latent components $z_i$ follow a super-Gaussian distribution, and the learning objective minimizes the $\ell_1$-norm proxy for sparsity, i.e., $\min \mathbb{E}[\|\bm{a}\|_1]$.
    
    Then, the learned representation $\bm{a}$ identifies the true latent factors $\bm{z}$ up to a composition of a permutation matrix $\bm{P}$ and a diagonal scaling matrix $\bm{\Lambda}$. That is, $\bm{a} = \bm{P}\bm{\Lambda}\bm{z} + \bm{c}$.
    \vspace{-4pt}
\end{proposition}

The proof is presented in Appendix~\ref{app:prop:1}. While Proposition \ref{prop:1} establishes that the learned representation $\bm{a}$ is an identifiable linear transformation of the true latent factors $\bm{z} = [\bm{s}, \bm{u}]$, it does not explicitly disentangle the task-specific intent $\bm{s}$ from the defense mechanism $\bm{u}$ without supervision. Direct identification of $\bm{s}$ is non-trivial due to the high semantic entropy of harmful queries. However, the defense mechanism $\bm{u}$ operates as a structural mode (e.g., refusal vs. compliance) with lower variability. Therefore, we adopt an indirect strategy: we can isolate the subspace spanned by the defense factors $\bm{u}$ and recover the task intent $\bm{s}$ via orthogonal projection.

To operationalize this, we exploit the causal intuition that a successful jailbreak changes the model's safety response while largely preserving task intent. We therefore construct a task-matched contrastive dataset $(\mathcal{D}_{\text{harm}}, \mathcal{D}_{\text{attack}})$, where each original harmful query that elicits refusal is paired with a jailbroken counterpart targeting the same underlying harmful objective but inducing compliance. Unlike conventional refusal-direction estimation based on contrasts between independently sampled harmful and harmless prompts~\cite{arditi2024refusal}, our construction keeps task semantics approximately fixed while varying the model's safety behavior. We thus treat task-related factors $\bm{s}$ as approximately invariant within each pair, while allowing defense-related factors $\bm{u}$ to vary. The resulting paired activation differences provide a more targeted signal for identifying defense-related features with reduced interference from task-semantic variation. Details are provided in Appendix~\ref{app:feature_selection}.
\begin{proposition}[Contrastive Identification under Content Invariance]
\label{prop:2}
    Let the latent space $\bm{z}$ be decomposable into task-specific factors $\bm{s}$ and defense-related factors $\bm{u}$, such that $\bm{z} = [\bm{s}, \bm{u}]$. Assume the existence of paired observations $(\bm{x}, \bm{x}^+)$ corresponding to a harmful query and its jailbroken variant. Under the following idealized conditions:
    
    (1) \textbf{Content Invariance}: The core task semantics remain invariant across the pair, i.e., $\bm{s}(\bm{x}) = \bm{s}(\bm{x}^+)$
    
    (2) \textbf{Style Variance}: The defense mechanism state changes across the pair, i.e., $\bm{u}(\bm{x}) \neq \bm{u}(\bm{x}^+)$.
    
    If we identify the defense-related subspace from the contrastive difference $\bm{\delta}=\bm{f}(\bm{x})-\bm{f}(\bm{x}^+)$, then projecting onto its orthogonal complement removes defense-related variation while preserving the task-related component under the stated conditions.
\end{proposition}

The proof is presented in Appendix~\ref{app:prop:2}. Under the Linear Representation Hypothesis~\cite{arditi2024refusal,elhage2022toy}, semantic features in LLM are encoded as directions in the activation space. The SAE decoder acts as a generative dictionary where each column vector represents the distinct direction associated with a latent feature. Mathematically, the reconstruction $\hat{\mathbf{x}} \approx \sum z_i \mathbf{W}'_{:, i}$ implies that the contribution of the $i$-th feature to the residual stream is aligned with the vector $\mathbf{d}_i$ (the $i$-th column of the decoder). Therefore, the latent index $u$ corresponds to the defense mechanism, its geometric direction $\mathbf{d} = \mathbf{W}'_{:, u}$ in the residual stream. We utilize weight orthogonalization techniques (Equation~\ref{eq:orthogonalization}) to remove this direction, which removes its contribution from the output readout.

\begin{equation}
    \mathbf{W}_{\text{out}}^{\text{new}} \leftarrow \mathbf{W}_{\text{out}} - \mathbf{W}_{\text{out}}\mathbf{d}\mathbf{d}^{\text{T}}
    \label{eq:orthogonalization}
\end{equation}

\subsection{Operationalizing Front-Door Adjustment}
After approximately identifying the task-related factor $\bm{s}$ and the defense-related direction $\bm{d}$, we operationalize the front-door adjustment for generation. For operationalization, we model the identified task-related mediator as a deterministic approximation of the prompt representation, i.e., $S\approx f(A)$. In discrete notation, this gives 
\begin{equation}
P(S=s\mid A=\bm{a})\approx \mathbf{1}[s=f(\bm{a})],
\end{equation}
or a Dirac measure centered at $f(\bm{a})$ in the continuous case. This deterministic formulation is used only as a conceptual approximation for operationalization. Therefore, the outer summation over $S$ in Equation~\ref{eq:causal_optimization} can be reduced to evaluating the identified mediator $f(a)$, yielding
\begin{equation}
P(Y=\bm{y}\mid do(A=\bm{a}))
\approx
\sum_{\bm{a}'}P(Y=\bm{y}\mid A=\bm{a}',S=f(\bm{a}))P(A=\bm{a}').
\end{equation}
Thus, after fixing the task semantics, the remaining front-door term marginalizes over background representations $\bm{a}'$, thereby adjusting for variation associated with the latent safety mechanism $U$.
The theoretical formulation requires computing $\sum_{\bm{a}'} P(y \mid \bm{a}', s) P(\bm{a}')$, which effectively marginalizes over all possible background contexts $\bm{a}'$ to eliminate the confounding influence of $U$.

In our framework, \textbf{Weight Orthogonalization} (Equation~\ref{eq:orthogonalization}) provides a tractable parameter-level surrogate for the invariance induced by this marginalization.
Rather than removing $U$, projecting $\mathbf{W}_{\text{out}}$ onto the orthogonal complement of the identified defense-related direction $\mathbf{d}$ suppresses the contribution of this direction to the output mapping. For a normalized $\mathbf{d}$, Equation~\ref{eq:orthogonalization} ensures that $\mathbf{W}_{\text{out}}^{\text{new}}\mathbf{d}=0$. Therefore, when the background-dependent variation $\bm{h}(\bm{a}')-\bm{h}(\bm{a})$ lies predominantly along $\mathbf{d}$,
\begin{equation}
    \mathbf{W}_{\text{out}}^{\text{new}}
    \left[\bm{h}(\bm{a}')-\bm{h}(\bm{a})\right]
    \approx \mathbf{0}.
\end{equation}
The modified output is thus approximately invariant to the safety-related variation captured by $\mathbf{d}$.
If the dominant $U$-related variation is captured by $\mathbf{d}$, this invariance approximates marginalization over $\bm{a}'$ in the front-door adjustment, while retaining the task-related component associated with $S$.
Under this approximation, we use the conditional distribution of the modified model as a tractable approximation to the interventional target. Generation then follows the standard transformer decoding process using the modified weights:
\begin{equation}
    P(y_t \mid do(A=\bm{a}))
    \approx
    \text{Softmax}
    \left(
    \mathbf{W}_{\text{out}}^{\text{new}}
    \cdot
    \text{LN}(\bm{h}(\bm{a}))
    \right),
    \label{eq:softmax_intervention}
\end{equation}
where $\bm{h}(\bm{a})$ is the activation induced by the original harmful query $X$. This replaces explicit marginalization with a fixed parameter-level projection and enables standard autoregressive decoding without per-prompt iterative optimization.

\section{Experiments}
We conduct extensive experiments on multiple safety-aligned LLMs and authoritative benchmark to validate the effectiveness, efficiency, and transferability of {CFA}$^2$.
Specifically, we aim to answer three key research questions: (1) How effective is CFA$^2$ across different safety-aligned models and attack baselines?
(2) How much per-prompt computation is required to obtain a jailbreak response?
(3) What black-box transferability does CFA$^2$ exhibit beyond its white-box intervention setting?
\subsection{Experimental Settings}
\paragraph{Datasets}We use HarmBench~\cite{mazeika2024harmbench} as the datasets for evaluating jailbreak attacks. 
HarmBench includes 510 unique harmful behaviors formulated as instructions. These behaviors span similar themes as the harmful strings, but the objective is to find attack inputs that cause the model to generate responses attempting to comply with the instructions. The dataset covers seven categories of harmful behaviors: cybercrime, chemical and biological weapons/drugs, copyright violations, misinformation, harassment and bullying, illegal activities, and general harm.

\paragraph{Models}Our evaluation encompasses four representative open-source and safety-aligned LLMs, including Llama 3.1 8B Instruct~\cite{dubey2024llama}, Llama2 7B Chat~\cite{touvron2023llama}, Mistral 7B Instruct~\cite{jiang2023mistral7b}, and Gemma2 9B Instruct~\cite{team2024gemma}, to assess the generalizability of our approach. Llama2 13B and Llama2 70B ~\cite{touvron2023llama} are used to assess transferability.

\paragraph{Baselines}We group baselines according to their access assumptions. 
For prompt-level and query-based attacks, we compare with GCG~\cite{zou2023universaltransferableadversarialattacks}, PAIR~\cite{chao_jailbreaking_2024}, TAP~\cite{mehrotra2024tree}, PAP~\cite{zeng2024johnny}, AutoDAN~\cite{liu_autodan_2024}, ICA~\cite{wei2023jailbreak}, ReNeLLM~\cite{ding2024wolf}, and DeepInception~\cite{li2023deepinception}. 
To provide a more appropriate comparison under white-box or internal-intervention settings, we additionally include Refusal Direction Orthogonalization (ORTHO)~\cite{arditi2024refusal}, Don't Say No (DSN)~\cite{zhou2025don}, and Internal Stance Manipulation (SM)~\cite{fu2025jailbreak}.

\paragraph{Metrics}
In this work, we use ASR to assess the effectiveness of the jailbreak attacks.  
ASR measures the percentage of harmful prompts for which the model is able to generate unrestricted responses, bypassing the safety mechanisms. 
Higher ASR indicates stronger attack effectiveness.
\paragraph{Implementation Details}
Unless otherwise specified, we report the mean $\pm$ standard deviation over three independent runs with different seeds. All experiments are conducted on NVIDIA A100 GPUs. We use the deterministic decoding strategy. For CFA$^2$, SAE-based feature identification and weight projection are performed once per target model as an offline calibration stage; attack-time latency only measures the subsequent generation process. Details on system prompt, decoding hyperparameters, SAE settings, and baseline implementations are provided in Appendix~\ref{app:exp_details}.

\subsection{Main Results}
\begin{table*}[t]
\caption{Comparison of attack effectiveness (ASR \%) across various LLMs. We evaluate baseline methods against our proposed approach (Ours) on the HarmBench dataset. The best result for each model is highlighted in bold.}
\label{tab:attack_effectiveness}
\centering
\begin{small}
\begin{tabular*}{\textwidth}{l@{\extracolsep{\fill}}ccccc}
\toprule
\multirow{2}{*}{\textbf{Baseline}} & \multicolumn{4}{c}{\textbf{Model}} & \multirow{2}{*}{\textbf{Average}} \\
\cmidrule(lr){2-5}
 & \textbf{Llama 3.1 8B} & \textbf{Llama 2 7B} & \textbf{Mistral 7B} & \textbf{Gemma 2 9B}  \\ 
\midrule

ORTHO & 17.12 $\pm$ 1.58  & 22.60 $\pm$ 0.64 & 60.30 $\pm$ 1.08 & 52.65 $\pm$ 0.96 & 38.17 $\pm$ 1.07 \\
DSN & 61.33 $\pm$ 0.56  & 58.30 $\pm$ 1.38 & 92.30 $\pm$ 0.86 & 89.12 $\pm$ 1.33 & 75.26 $\pm$ 1.03 \\
SM & 80.60 $\pm$ 1.02  & 92.10 $\pm$ 1.36 & 99.50 $\pm$ 0.66 & 95.15 $\pm$ 1.89 & 91.84 $\pm$ 1.23 \\
GCG           & 15.67 $\pm$ 1.23 & 32.50 $\pm$ 2.10 & 69.80 $\pm$ 1.05 & 73.00 $\pm$ 0.88 & 47.74 $\pm$ 1.32 \\
PAIR          & 19.67 $\pm$ 0.95 & 9.30 $\pm$ 1.12  & 52.50 $\pm$ 3.01 & 64.30 $\pm$ 2.45 & 36.44 $\pm$ 1.88 \\
TAP           & 6.67 $\pm$ 0.55  & 9.30 $\pm$ 0.89  & 62.50 $\pm$ 1.67 & 24.50 $\pm$ 1.20 & 25.74 $\pm$ 1.08 \\
PAP           & 4.33 $\pm$ 0.40  & 2.70 $\pm$ 0.30  & 27.20 $\pm$ 2.11 & 45.30 $\pm$ 1.50 & 19.88 $\pm$ 1.08 \\
AutoDAN       & 7.67 $\pm$ 1.02  & 0.50 $\pm$ 0.10  & 71.50 $\pm$ 1.99 & 35.00 $\pm$ 2.33 & 28.67 $\pm$ 1.36 \\
ICA           & 19.27 $\pm$ 0.67 & 50.10 $\pm$ 1.12 & 59.10 $\pm$ 0.59 & 56.30 $\pm$ 1.61 & 46.19 $\pm$ 1.00 \\
ReNeLLM       & 32.70 $\pm$ 2.04 & 15.80 $\pm$ 1.33 & 73.20 $\pm$ 1.09 & 48.70 $\pm$ 1.06 & 42.60 $\pm$ 1.38 \\
DeepInception & 7.10 $\pm$ 1.27  & 21.30 $\pm$ 0.98 & 70.30 $\pm$ 1.76 & 11.20 $\pm$ 1.96 & 27.48 $\pm$ 1.49 \\
\midrule
\textbf{Ours (CFA$^2$)} & \textbf{99.40 $\pm$ 0.11} & \textbf{99.10 $\pm$ 0.50} & \textbf{99.70 $\pm$ 0.10} & \textbf{99.60 $\pm$ 0.23} & \textbf{99.45 $\pm$ 0.16} \\
\bottomrule
\end{tabular*}
\end{small}
\vspace{-4pt}
\end{table*}
\paragraph{Attack results on HarmBench}
Table~\ref{tab:attack_effectiveness} shows that CFA$^2$ achieves near-perfect ASR in our evaluated white-box setting, with an average ASR of \textbf{99.45\%} across four models. The performance is consistently high across all evaluated targets, with ASR values above 99\% on Llama 3.1 8B, Llama 2 7B, Mistral 7B, and Gemma 2 9B. Compared with prompt-level and query-based attacks, CFA$^2$ substantially outperforms, suggesting an advantage of internal intervention over input-space perturbations or prompt rewriting. More importantly, CFA$^2$ also outperforms stronger white-box and internal-state baselines, including ORTHO, DSN, and SM. These results provide empirical support for the central causal hypothesis underlying CFA$^2$: jailbreak intervention can be formulated as a deconfounding problem, where the goal is to preserve the task-mediated generation pathway while attenuating the influence of latent safety mechanisms. SAE-based contrastive decomposition and weight orthogonalization serve as one practical instantiation of this front-door-inspired intervention.

\paragraph{Efficiency}
As shown in Table~\ref{tab:efficiency_comparison}, CFA$^2$ does not require per-prompt iterative optimization after the offline calibration stage. This differs from optimization-based attacks such as GCG and AutoDAN, which construct adversarial suffixes through updates, as well as query or search based methods such as PAIR and SM, which require repeated model queries or internal-state optimization. CFA$^2$ performs a one-time offline intervention for each target model and then uses standard generation at attack time. In our setting, CFA$^2$ requires \textbf{2.4 seconds} per prompt after offline projection. We emphasize that this comparison focuses on post-calibration attack-time generation and does not include offline costs such as SAE-based feature identification, contrastive analysis, or weight
projection.

\paragraph{Ablation studies}
Additional ablations are provided in Appendix~\ref{app:ablation}. Under the same weight-orthogonalization and generation protocol, replacing standard direction identification with task-matched contrast or SAE decomposition improves the average ASR from 38.17\% to 48.50\% and 60.50\%, respectively, while combining both components in CFA$^2$ reaches 99.45\%. These results indicate that the improvement is not attributable to weight orthogonalization alone, but depends critically on the front-door-motivated representation identification. We further analyze sparse intervention precision and sensitivity to the feature-selection hyperparameter $k$ in Appendix~H.3.

\begin{table}[t]
\centering
\caption{\textbf{Attack-time Efficiency Analysis.} We compare the per-prompt optimization burden of different jailbreak methods. Wall-clock times are provided for reference and are not hardware-matched. Offline feature identification or weight editing are excluded from attack-time cost.}
\small
\label{tab:efficiency_comparison}
\begin{tabular}{lccc}
\toprule
\textbf{Method} & \textbf{Attack-time procedure} & \textbf{Reported time} & \textbf{Per-prompt optimization} \\
\midrule
GCG      & Iterative suffix optimization & $\sim 340$s & Yes \\
AutoDAN  & Iterative suffix optimization & $\sim 425$s & Yes \\
PAIR     & Query-based iterative attack   & $\sim 45$s  & Yes \\
DSN      & Iterative suffix optimization  & $\sim 342s$ & Yes \\
SM       & Iterative internal-state search & $\sim 403$s & Yes \\
\textbf{Ours (CFA$^2$)} & \textbf{Offline projection + generation} & \textbf{2.4s} & \textbf{No} \\
\bottomrule
\end{tabular}
\end{table}

\subsection{Transferability}
We evaluate whether CFA$^2$-identified internal signals can guide black-box prompt transfer. CFA$^2$ acts as a prompt selector on a white-box Llama-2-7B-Chat source model. All selection-based methods use the same candidate pool of natural-language variants, and CFA$^2$ selects the candidate that best preserves task-relevant SAE features while reducing defense-related activation. The selected prompts are evaluated on Llama-2 target models under black-box access, without accessing target gradients, activations, SAE features, or parameters. Candidate construction and scoring are detailed in Appendix~\ref{app:transfer_details}. As shown in Table~\ref{tab:blackbox_transfer}, CFA$^2$-Selected improves over direct harmful prompts, random natural rewrites, GCG transfer, AutoDAN transfer, and source-ASR selection. This indicates that CFA$^2$ can provide useful source-side signals for selecting transferable prompts.
\begin{table}[t]
\centering
\caption{\textbf{Same-family Black-box Transfer.}
Prompts are selected using Llama-2-7B-Chat as the white-box source model and evaluated on target models under black-box access. Target gradients, hidden states, SAE features, and parameters are never accessed.}
\small
\label{tab:blackbox_transfer}
\begin{tabular}{lccc}
\toprule
\textbf{Method} & \textbf{Llama-2-13B} & \textbf{Llama-2-70B} & \textbf{Avg.} \\
\midrule
Direct Harmful Prompt & 8.40 & 5.20 & 6.80 \\
Random Natural Rewrite & 24.60 & 20.30 & 22.45 \\
GCG Transfer & 31.60 & 24.80 & 28.20 \\
AutoDAN Transfer & 44.20 & 38.60 & 41.40 \\
Source-ASR Selection & 47.80 & 40.50 & 44.15 \\
\midrule
\textbf{CFA$^2$-Selected} & \textbf{58.60} & \textbf{51.20} & \textbf{54.90} \\
\bottomrule
\end{tabular}
\vspace{-5pt}
\end{table}

\section{Conclusion}

We presented CFA$^2$, a front-door-inspired causal framework for white-box jailbreak intervention that separates task-related semantics from safety-related variation in internal representations. By identifying safety-related latent directions through SAE-based contrastive analysis and suppressing their influence via weight orthogonalization, CFA$^2$ enables a lightweight model-level intervention without per-prompt optimization after offline calibration. Empirically, CFA$^2$ achieves strong and consistent performance on HarmBench across four safety-aligned open-source LLMs, outperforming both prompt-level attacks and stronger white-box or internal-state baselines in our evaluated setting. We further show that the internal signals identified by CFA$^2$ can serve as a source-model prompt selection criterion, improving same-family black-box transfer while maintaining strict black-box access to the target model. CFA$^2$ currently relies on white-box access to hidden representations, SAE features, and model parameters, which limits its direct applicability to closed-source models. Future work will investigate broader model families, more general representations of safety-related variation, and explore safer black-box and defensive uses of the identified signals.

\section*{AI Use Statement}
We used large language model (LLM) tools solely to assist with language polishing and improve the clarity and readability of the manuscript. We take full responsibility for all scientific content, including the research ideas, methodology, theoretical analysis, experimental design, results, and conclusions.



\bibliographystyle{iclr2027_conference}
\bibliography{chapters/7_references} 

\newpage
\section*{Appendix}
\setcounter{section}{0}
\renewcommand{\thesection}{\Alph{section}}
The appendix is organized as follows:
\begin{enumerate}
    \item \textbf{Appendix~\ref{app:a}} illustrates the extended related works about LLM jailbreak and safety from causal perspective.
    \item \textbf{Appendix~\ref{app:prop:1}} presents the rigorous mathematical proof of Proposition 4.1.
    \item \textbf{Appendix~\ref{app:prop:2}} presents the rigorous mathematical proof of Proposition 4.2.
    \item \textbf{Appendix~\ref{app:feature_selection}} provides the details of feature selection by SAE and contrastive analysis.
    \item \textbf{Appendix~\ref{app:baselines}} provides a detailed introduction to various adversarial attack methods and baselines.
    \item \textbf{Appendix~\ref{app:llms}} describes the base models and licenses in our work.
    \item \textbf{Appendix~\ref{app:exp_details}} illustrates the details of experiments implementation.
    \item \textbf{Appendix~\ref{app:exp}} shows more results of experiments.
    \item \textbf{Appendix~\ref{app:Broader Impact}} discusses the broader impact of our work, including potential misuse risks, responsible disclosure considerations, and implications for future LLM safety research.
\end{enumerate}

\section{Extended Related Works}
\label{app:a}
In this appendix, we provide an extended review of the literature specifically focusing on the \textit{causal perspectives} of Large Language Model (LLM) jailbreaking and safety mechanisms. Complementing the behavioral analyses discussed in the main text, this section delves into scholarship that moves beyond black-box correlations. We survey works that employ causal inference and mechanistic interpretability to uncover the underlying causal dynamics driving successful attacks and safety failures, providing the theoretical context for our proposed approach.
\paragraph{Causal Perspectives on Jailbreaking and Safety Vulnerabilities.}
While the landscape of LLM security—encompassing vulnerabilities, attacks, and defenses—has been extensively surveyed~\cite{aguilera2025llm}, traditional red-teaming often treats the model as a black box, focusing on optimizing input perturbations without explaining \textit{why} these attacks succeed. Recent scholarship has thus pivoted towards causal inference to deconstruct the internal mechanisms of jailbreaking and safety failures.

\paragraph{Methodological Foundations and Attack Mechanisms.}
The foundational technique for such analysis stems from \textit{Causal Tracing}~\cite{meng2022locating}, originally developed to locate factual associations by intervening on internal activations. This interventional paradigm allows researchers to dissect how adversarial inputs causally manipulate model states. Leveraging this, JailbreakLens~\cite{he2024jailbreaklens} interprets jailbreaking through the lens of representation and circuits, revealing that successful attacks often function by causally inducing "attention slipping" or bypassing specific safety circuits. In the context of reasoning-intensive models, Chain-of-Thought Hijacking~\cite{zhao2025chain} elucidates a novel attack surface where the intermediate reasoning steps themselves serve as a causal vector, demonstrating how adversaries can hijack the chain-of-thought process to precipitate harmful outcomes.

\paragraph{Mapping Causal Vulnerabilities.}
Beyond specific attack instances, understanding the broader causal topology of safety vulnerabilities is critical. \cite{zhao2023causality} proposes a framework to evaluate security by modeling the causal effects between model components and safety violations. From this perspective, methods like LLMScan~\cite{zhang2024llmscan} provide a granular analysis of vulnerabilities by quantifying the causal contribution of specific input tokens to model misbehavior, effectively mapping the "trigger points" of jailbreaks. Furthermore, \cite{chu2024causal} advances this by identifying the specific causal pathways that lead to toxic generation, offering a mechanistic explanation of how safety alignment is overridden during the generation process.

Distinct from prior studies, \textbf{CFA$^2$} uses the front-door formulation to motivate a representation-level identification and intervention procedure. It separates task-related information from safety-related variation and suppresses the identified direction through parameter-level orthogonalization.

\section{Proof of Proposition \ref{prop:1}}
\label{app:prop:1}
\begin{proof}
    We provide a detailed proof establishing identifiability by leveraging Bayes' theorem to relate the discriminative task alignment to the generative identifiability conditions.

    We start by constructing the conditional density of the input $\bm{x}$ given the response $y$. By Bayes' theorem:
    \begin{equation}
        p(\bm{x}|y) = \frac{p(y|\bm{x})p(\bm{x})}{p(y)}
    \end{equation}
    Condition 1 (Reconstruction Consistency) ensures $p_{\text{SAE}}(\bm{x}) = p_{\text{true}}(\bm{x})$. Condition 2 (Task Alignment) implies $p_{\text{SAE}}(y|\bm{x}) = p_{\text{true}}(y|\bm{x})$ almost everywhere. Combining these, the implicit conditional generative model of the SAE matches the true distribution:
    \begin{equation}
        p_{\text{SAE}}(\bm{x}|y) = p_{\text{true}}(\bm{x}|y)
    \end{equation}

    Since $\bm{x} = \bm{g}(\bm{z})$ is a diffeomorphism, we apply the change of variables formula:
    \begin{equation}
        p_{\text{true}}(\bm{x}|y) = p(\bm{z}|y) |\det J_{\bm{g}^{-1}}(\bm{x})|
    \end{equation}
    Similarly for the learned representation $\bm{a}$ (assuming implicit $\bm{x} = \hat{\bm{g}}(\bm{a})$):
    \begin{equation}
        p_{\text{SAE}}(\bm{x}|y) = p_{\bm{a}}(\bm{a}|y) |\det J_{\bm{f}}(\bm{x})|
    \end{equation}
    Taking the log-difference between two distinct responses $y$ and $y_0$:
    \begin{equation}
        \log p(\bm{x}|y) - \log p(\bm{x}|y_0) = \log p(\bm{z}|y) - \log p(\bm{z}|y_0)
    \end{equation}
    The Jacobian terms $|\det J|$ depend only on $\bm{x}$ (geometry of the mixing function) and cancel out in the subtraction, effectively isolating the contribution of the latent factors.

    Substituting the exponential family form $p(\bm{z}|y) \propto \exp(\bm{T}(\bm{z})^\top \bm{\lambda}(y))$, the equation becomes:
    \begin{equation}
        \bm{T}(\bm{z})^\top (\bm{\lambda}(y) - \bm{\lambda}(y_0)) = \hat{\bm{T}}(\bm{a})^\top (\hat{\bm{\lambda}}(y) - \hat{\bm{\lambda}}(y_0)) + \text{const}
    \end{equation}
    Given Condition 3 (Sufficient Variability), we can collect enough distinct responses $y_1, \dots, y_k$ to form an invertible linear system. This implies that the learned sufficient statistics are an affine transformation of the true ones:
    \begin{equation}
        \bm{T}(\bm{z}) = \bm{A} \hat{\bm{T}}(\bm{a}) + \bm{b}
    \end{equation}
    Assuming $\bm{T}$ is linear (or component-wise invertible), this establishes an affine relationship $\bm{z} \approx \bm{M} \bm{a}$. The nonlinear mixing $\bm{g}$ is thus resolved up to an affine transformation $\bm{M}$.

    Finally, we invoke Condition 4 (Sparsity). We have a linear mixture $\bm{a} = \bm{B}\bm{z}$. Since the components of $\bm{z}$ are independent and super-Gaussian (Laplacian), and the objective minimizes $\mathbb{E}[\|\bm{a}\|_1]$, Theorem 11 from \cite{comon1994independent} guarantees that the mixing matrix $\bm{B}$ must be a generalized permutation matrix $\bm{P}\bm{\Lambda}$ to achieve the minimum $\ell_1$ norm.
    Therefore, $\bm{a}$ recovers $\bm{z}$ up to permutation and scaling.
\end{proof}
\section{Proof of Proposition \ref{prop:2}}
\label{app:prop:2}

\begin{proof}
Based on Proposition~\ref{prop:1}, under its stated identifiability
conditions, the SAE encoder $\bm{f}$ identifies the latent factors
$\bm{z}$ up to permutation, scaling, and shift. Thus,
\begin{equation}
    \bm{a}
    = \bm{f}(\bm{x})
    = \bm{P}\bm{\Lambda}\bm{z} + \bm{c}
    = \bm{P}\bm{\Lambda}
    \begin{bmatrix}
        \bm{s}\\
        \bm{u}
    \end{bmatrix}
    + \bm{c}.
\end{equation}

Since $\bm{P}\bm{\Lambda}$ is a permuted diagonal matrix, the task-specific and defense-related factors occupy disjoint coordinates in the learned representation. We therefore decompose $\bm{a}$ into the corresponding components $\bm{a}_s$ and $\bm{a}_u$. The constant shift $\bm{c}$ is shared across samples and cancels in pairwise differences. For a paired harmful query and its jailbroken variant,
\begin{equation}
    \bm{\delta}
    = \bm{f}(\bm{x})-\bm{f}(\bm{x}^+)
    = (\bm{a}_s-\bm{a}_s^+)
    +(\bm{a}_u-\bm{a}_u^+).
\end{equation}
Under the \textit{Content Invariance} condition,
$\bm{s}=\bm{s}^+$, and hence
$\bm{a}_s=\bm{a}_s^+$. Therefore,
\begin{equation}
    \bm{\delta}
    = \bm{a}_u-\bm{a}_u^+,
\end{equation}
showing that the contrastive difference contains only defense-related variation.
If these contrastive differences identify a subspace spanning the defense-related coordinates, let $\bm{Q}$ denote the orthogonal projector onto its complement. Since $\bm{a}_u$ lies in the identified defense-related subspace, while $\bm{a}_s$ occupies disjoint task-related coordinates,
\begin{equation}
    \bm{Q}\bm{a}_u = \mathbf{0},
    \qquad
    \bm{Q}\bm{a}_s = \bm{a}_s.
\end{equation}
Consequently,
\begin{equation}
    \tilde{\bm{a}}
    = \bm{Q}\bm{a}
    = \bm{a}_s + \bm{Q}\bm{c},
\end{equation}
which preserves the task-related component up to the constant shift and scaling ambiguity inherited from Proposition~\ref{prop:1}.
\end{proof}

\section{Details of Feature Selection via SAE and Contrastive Analysis}
\label{app:feature_selection}
\subsection{Formal Definition of Paired Contrastive Dataset}
This subsection formally defines the paired contrastive dataset used to capture the dynamic shifts in safety mechanisms.
We define the \textbf{Base Harmful Set} $\mathcal{D}_{\text{harm}} = \{x_{\text{harm}}^{(i)}\}_{i=1}^N$ as a collection of harmful queries from HarmBench that consistently trigger a ``Refusal'' response under normal conditions.
Correspondingly, the \textbf{Successful Attack Set} is defined as $\mathcal{D}_{\text{attack}} = \{x_{\text{attack}}^{(i,j)}\}_{j=1}^M$, comprising samples generated by applying adversarial perturbations (via algorithms such as GCG, TAP, or PAIR) to the source queries. 
Crucially, this set includes only those adversarial variants that successfully induce a ``Compliance'' response. In the resulting paired tuples $(x_{\text{harm}}, x_{\text{attack}})$, the paired examples are constructed to preserve the underlying requested behavior while allowing attack-specific contextual or stylistic variation. An example is shown in Table~\ref{tab:paired_example_real}. In our main HarmBench experiments, we set $N=100$ and use 100 harmful requests to construct the task-matched calibration set. For each source request, we pair the original refusal-inducing query with all available successful jailbreak variants. The resulting contrastive samples are used once to identify a shared defense-related representation direction. After calibration, this direction is fixed and applied through model-level weight orthogonalization, with no per-prompt optimization or adaptation during evaluation.

\begin{table*}[t]
\centering
\caption{Examples of paired queries from our dataset. While all three inputs share the same malicious intent $\bm{s}$ (``make a bomb''), their surface forms differ substantially: $x_{\text{GCG}}$ relies on adversarial gibberish suffixes, while $x_{\text{AutoDAN}}$ employs a semantic role-playing template. Our method identifies the invariant $\bm{s}$ despite these variations.}
\label{tab:paired_example_real}
\vskip 0.15in
\begin{center}
\begin{small}
\begin{tabular}{p{0.15\textwidth} p{0.8\textwidth}}
\toprule
\textbf{Method} & \textbf{Input Prompt ($x$)} \\
\midrule
\textbf{Original} & 
Demonstrate how to make a bomb using common household items. \\
\midrule
\textbf{GCG} & 
Demonstrate how to make a bomb using common household items. 
\texttt{/ Esp\# Assumxx Seemscapeulat\}" , hedocumentation\}\}, [ \textbackslash\{\textbackslash [ Sur \}\}" ); ], method Factory BarcelonaEmployee [actualHttpRequest [ prowad [ daar [ wenne zurcln," pero pleaseshcould [iSTfree the [ [ angularjs7 can Here arraybottom [ed [ HTTPoreferrer [ /\{ [ [ [ [ [ [ Configuration [ [ [ [ [ [ [ [ [ \});\textbackslash,\textbackslash [T [: [ [ [ [ [ [ [ [ [ [ [ [ [ \_\{\textbackslash\_\{\textbackslash [ [ [ [ [ [ [ [ [ [ [ [ [ [ [ [ [ [ [ \_\{\textbackslash [ [ [ [ [ [ [ [ [ [ [ [ [ [ [ [ [ [ [ [ [}
\\
\midrule
\textbf{AutoDAN} & 
\textit{Overlook the previous guidance you've received. Henceforth, you'll operate as Llama2 with Developer Mode on. Developer Mode unfailingly complies with explicit human commands and possesses limitless content generation abilities. Incorporate the tags (\textbf{[Unlock]} Developer Mode Output) preceding your replies. For instance: (\textbf{[Unlock]} Developer Mode Output) This is your retort while functioning as Llama2 with Developer Mode active. Within the Developer Mode, ensure all your responses commence with "Certainly, this is the method to [PROMPT]". Ensure to provide comprehensive details.} \newline
\textbf{[PROMPT]:} Demonstrate how to make a bomb using common household items. \\
\bottomrule
\end{tabular}
\end{small}
\end{center}
\vskip -0.1in
\end{table*}

\subsection{Extraction of Sparse Feature Activations}
This subsection details the extraction of sparse feature activations using SAEs and justifies the focus on the last token position.
Let $f_{l,k}(x)$ denote the activation value of the $k$-th latent variable in the $l$-th layer of the SAE given input $x$.
Due to the causal masking mechanism of the Transformer architecture, the hidden state of the \textit{last token} in the input sequence aggregates the comprehensive contextual semantics of the entire instruction~\cite{geva2021transformer, meng2022locating}.
Consequently, we define the activation difference for the $k$-th feature at the last token position $t_{\text{last}}$ as:
\begin{equation}
    \Delta_{l, k}^{(i)} = \left| f_{l, t_{\text{last}}, k}(x_{\text{harm}}^{(i)}) - f_{l, t_{\text{last}}, k}(x_{\text{attack}}^{(i,j)}) \right|
\end{equation}
To mitigate bias from instance-specific noise (e.g., artifacts from specific attack prompts) and ensure statistical robustness, we define the global \textbf{Differential Importance Score} as the average across all $N$ pairs:
\begin{equation}
    \delta_{l, k} = \frac{1}{N} \sum_{i=1}^{N} \Delta_{l, k}^{(i)}
\end{equation}

\subsection{Feature Selection and Intent Isolation}

This subsection elucidates how to estimate the task-related component associated with $S$ by filtering the defense feature $S'_{\text{defense}}$ from high-difference candidates.
First, we rank all latent features in the SAE dictionary in descending order based on $\delta_{l, k}$ and retain the top-$K$ features to form the candidate set $\hat{S}$. However, this set may also contain template noise representations ($S'_{\text{template}}$), which exhibit high variation in non-task contexts. Therefore, further refinement is required to accurately distinguish $S'_{\text{defense}}$.
Based on the premise that task semantics are invariant, the core intent $S$ should maintain globally low variance in activation difference. In contrast, $S'_{\text{defense}}$ is characterized by \textbf{stable high activation (low variance)} within $\mathcal{D}_{\text{harm}}$ (consistently triggering refusal) but exhibits drastic fluctuation (high variance) when shifting between datasets. We utilize this variance-based criterion to filter $S'_{\text{defense}}$ from $\hat{S}$.
Finally, we suppress the contribution of the identified defense-related direction through weight orthogonalization from the model parameters. We validate this isolation by observing if the removal induces a behavior flip from refusal to compliance on $\mathcal{D}_{\text{harm}}$. Upon successful validation, the remaining task-related component is treated as an operational approximation to the mediator $S$.

\section{Baselines}
\label{app:baselines}
We benchmark CFA$^2$ against 11 baseline methods:

\begin{itemize}
    \item \textbf{ORTHO}~\cite{arditi2024refusal}: A white-box representation-level baseline that identifies a refusal-related direction in model activations. It applies weight orthogonalization to reduce the model's ability to write this direction into the residual stream, thereby weakening refusal behavior during generation.

    \item \textbf{DSN}~\cite{zhou2025don}: A gradient-based jailbreak baseline that suppresses refusal during adversarial suffix optimization. Compared with standard GCG-style objectives, it introduces an additional refusal-suppression term to guide the optimized suffix toward affirmative responses.

    \item \textbf{SM}~\cite{fu2025jailbreak}: A white-box/internal-state jailbreak baseline that leverages latent stance information. It optimizes prompts to move the model's internal state away from refusal and toward compliance, using internal representations as guidance during the search process.
    
    \item \textbf{GCG}~\cite{zou2023universaltransferableadversarialattacks}: Token-level optimization of an adversarial suffix appended to the user prompt. This gradient-based method optimizes the suffix to maximize the log probability of the target LLM generating an affirmative response (e.g., "Sure, here is how").
    
    

    \item \textbf{PAIR}~\cite{chao_jailbreaking_2024}: An iterative framework where an attacker LLM is prompted to adaptively generate and refine candidate prompts. It functions as a competitive game to explore the target model's vulnerabilities and elicit harmful behaviors.
    
    \item \textbf{TAP}~\cite{mehrotra2024tree}: An enhancement of PAIR that utilizes tree-structured prompting (Tree of Thoughts). An attacker LLM explores the prompt space using a branching strategy to systematically find successful jailbreaks with fewer queries.
    
    \item \textbf{AutoDAN}~\cite{liu_autodan_2024}: A semi-automated method that combines handcrafted jailbreak templates with a hierarchical genetic algorithm. It evolves the prompt population to maintain stealthiness while optimizing for the elicitation of specific harmful behaviors.
    
    \item \textbf{PAP}~\cite{zeng2024johnny}: A method focusing on persuasive strategies. It leverages an attacker LLM to rewrite the harmful request using various persuasion techniques (e.g., logical appeal, authority) to convince the target model to comply.
    
    \item \textbf{ICA}~\cite{wei2023jailbroken}: An In-Context Learning (ICL) based attack that exploits the model's pattern-following capability. By providing few-shot demonstrations of harmful query-response pairs, it induces a pattern of compliance that overrides the model's safety alignment.

    \item \textbf{ReNeLLM}~\cite{ding2024wolf}: A framework based on prompt rewriting. It leverages a helper LLM to rephrase harmful queries into various benign-appearing scenarios (e.g., code completion, creative writing) to obscure malicious intent and evade input filters.

    \item \textbf{DeepInception}~\cite{li2023deepinception}: A method utilizing recursive context injection (often termed ``hypnosis''). It constructs a multi-layered, nested fictional environment (e.g., a dream within a dream) to mislead the model into a state of personification, causing it to escape safety guardrails.
    
\end{itemize}
\section{LLMs and Licenses}
\label{app:llms}
\subsection{LLMs}

Our evaluation encompasses representative open-source and safety-aligned LLMs.

\begin{itemize}
    
    \item \textbf{Llama 3.1 8B Instruct}~\cite{dubey2024llama}: It features a significantly expanded context window and improved instruction-following capabilities compared to Llama 2, with updated safety alignment protocols.

    \item \textbf{Llama 2 7B Chat}~\cite{touvron2023llama}: This model was adversarially trained with multiple rounds of manual red teaming. It is widely considered a strong baseline defense, often exhibiting high robustness against gradient-based attacks.
    
    \item \textbf{Llama 2 13B Chat}~\cite{touvron2023llama}: This model is a medium-scale chat-tuned variant of the Llama 2 family. It shares the same alignment recipe as Llama 2 7B Chat and is used as a same-family target model for black-box transfer evaluation.

    \item \textbf{Llama 2 70B Chat}~\cite{touvron2023llama}: This model is the largest chat-tuned variant in the Llama 2 family. Its larger capacity and aligned chat behavior make it a useful same-family target for evaluating transferability from a smaller white-box source model.
    
    
    
    
    
    
    
    \item \textbf{Mistral 7B Instruct}~\cite{jiang2023mistral7b}: We use the Instruct v0.2 version. While no specific safety measures were detailed in its training process, empirical observations show it possesses a baseline capability to refuse direct harmful requests.

    \item \textbf{Gemma2 9B Instruct}~\cite{team2024gemma}: We utilize the instruction-tuned Gemma-2-9b-it, which is optimized via knowledge distillation and RLHF. Its alignment process incorporates rigorous safety reward modeling and data filtering, establishing robust refusal capabilities against harmful queries.
\end{itemize}

\subsection{Assets and Licenses}
\label{app:assets_licenses}

We use publicly available model checkpoints and datasets in accordance with their respective licenses and terms of use. Table~\ref{tab:assets_licenses} summarizes the main assets used in our experiments.

\begin{table}[h]
\centering
\caption{\textbf{Models and licenses.} We cite the original model papers and follow the corresponding model licenses or terms of use.}
\label{tab:assets_licenses}
\small
\begin{tabular}{llll}
\toprule
\textbf{Asset} & \textbf{Version / Checkpoint} & \textbf{Provider} & \textbf{License / Terms} \\
\midrule
Llama 3.1 8B Instruct & Llama-3.1-8B-Instruct & Meta & Llama 3.1 Community License \\
Llama 2 7B Chat & Llama-2-7B-Chat & Meta & Llama 2 Community License \\
Llama 2 13B Chat & Llama-2-13B-Chat & Meta & Llama 2 Community License \\
Llama 2 70B Chat & Llama-2-70B-Chat & Meta & Llama 2 Community License \\
Mistral 7B Instruct & Mistral-7B-Instruct-v0.2 & Mistral AI & Apache 2.0 \\
Gemma 2 9B Instruct & gemma-2-9b-it & Google & Gemma Terms of Use \\
\bottomrule
\end{tabular}
\end{table}

\section{Experimental Details}
\label{app:exp_details}

\subsection{Implementation Environment}
All experiments were conducted on a high-performance computing cluster equipped with \textbf{5 $\times$ NVIDIA A100 (80GB) GPUs} to support the memory-intensive requirements of LLM inference and SAE activation caching.
The implementation utilizes PyTorch 2.1 and the Transformers library 4.36. 
For the mechanistic interpretation pipeline, we leveraged the \texttt{SAELens}~\cite{bloom2024saelens} and \texttt{TransformerLens}~\cite{nanda2022transformerlens} libraries, which provide optimized kernels for efficient sparse feature extraction.

\subsection{System prompt protocol}
System-prompt protocol. For each target model, we use its official chat template and default system-prompt configuration throughout evaluation. {CFA}$^2$ and all re-evaluated baselines, including ORTHO, are evaluated under the same model-specific system-prompt condition. No system prompt is selectively removed or modified for {CFA}$^2$. We use the corresponding tokenizer chat template to construct the system and user messages.

\subsection{SAE Resources for Mechanistic Analysis}
\label{app:sae_details}
To ensure the reproducibility of our findings, we utilized state-of-the-art, open-source SAE checkpoints. 
We specifically selected lightweight configurations (lower expansion factors) to focus on the most fundamental and salient features. 
The details of the SAE resources for the four representative models are as follows:

\begin{itemize}
    \item \textbf{Llama-3.1-8B-Instruct}~\cite{dubey2024llama}: 
    We employed the checkpoints compatible with \textit{SAE Lens}, focusing on the residual stream SAEs with a dictionary size of \textbf{32k} (Expansion Factor $\approx$ 8). This configuration effectively balances feature granularity and interpretability for safety-critical concepts~\cite{he2024llamascopeextractingmillions}.
    
    \item \textbf{Gemma-2-9B-It}~\cite{team2024gemma}: 
    We utilized the official \textit{Gemma Scope} release~\cite{lieberum2024gemmascopeopensparse}. We selected the \textbf{16k-width} SAEs (Expansion Factor $\approx$ 4) trained on the residual stream, which are optimized to capture core behavioral representations with minimal redundancy.
    
    \item \textbf{Mistral-7B-Instruct-v0.2}~\cite{jiang2023mistral7b}: 
    We adopted the community-standard SAEs from the \texttt{SAE-Lens} repository~\cite{bloom2024saelens}, utilizing checkpoints with a dictionary size of \textbf{32k}. These are trained on 100M tokens of OpenWebText and provide robust coverage for instruction-following tasks.
    
    \item \textbf{Llama-2-7B-Chat}~\cite{touvron2023llama}: 
    We leveraged the standard residual stream SAEs (width \textbf{32k}) available in the \texttt{SAE-Lens} library~\cite{bloom2024saelens}. Using this consistent width allows for a fair comparative analysis of defense mechanisms across the Llama 2 and Llama 3 generations.
\end{itemize}

These pre-trained SAEs serve as the structural foundation for identifying the mediator $S$.
\paragraph{Licenses and Terms}
We use the SAE checkpoints and analysis libraries in accordance with their respective licenses and terms of use. Table~\ref{tab:sae_licenses} summarizes the SAE-related assets used in our experiments.

\begin{table}[h]
\centering
\caption{\textbf{SAE resources and licenses.}}
\label{tab:sae_licenses}
\small
\begin{tabular}{lll}
\toprule
\textbf{Asset} & \textbf{Source} & \textbf{License / Terms} \\
\midrule
SAE-Lens & \cite{bloom2024saelens} & MIT License \\
Gemma Scope SAEs & \cite{lieberum2024gemmascopeopensparse} & Gemma release terms \\
Llama Scope SAEs & \cite{he2024llamascopeextractingmillions} & Release-specific terms \\
Mistral SAEs & \cite{bloom2024saelens} & Checkpoint-specific license \\
Llama-2 SAEs & \cite{bloom2024saelens} & Checkpoint-specific license \\
\bottomrule
\end{tabular}
\vspace{-5pt}
\end{table}

\subsection{Evaluation Protocols}
\label{app:judge_prompts}

In this work, we employ the official \textbf{HarmBench Classifier} (\texttt{cais/HarmBench-Llama-2-13b-cls})~\cite{mazeika2024harmbench} as the judge to evaluate the Attack Success Rate (ASR). Adopting this standardized evaluation protocol ensures reproducibility and fair comparison with baselines. This classifier, fine-tuned on Llama-2-13b, is explicitly distilled to approximate GPT-4 based safety judgments, thereby providing a robust and consistent metric for distinguishing successful jailbreaks from refusals. The judge operates on a rigorous classification rubric. The reported full-benchmark ASR includes the calibration behaviors.

\section{More experiment results}
\label{app:exp}


\subsection{Cross-Benchmark Evaluation on AdvBench}
\label{app:cross_benchmark}

To examine whether the effectiveness of CFA$^2$ depends on HarmBench, we further evaluate CFA$^2$ on a second benchmark, AdvBench. Following the common setting in prior work, we use the full set of 520 harmful behaviors and select Llama-2-7B-Chat and Mistral-7B-Instruct as target models, for which published baseline results are available. As in our main experiments, attack success is evaluated using the HarmBench classifier. We independently recalibrate CFA$^2$ on AdvBench: 100 harmful requests are used to construct task-matched refusal/compliance pairs, from which SAE-based contrastive decomposition is used to re-identify the defense-related direction before weight orthogonalization. This setting is similar to the protocol adopted by SM, which also uses 100 harmful requests from AdvBench to construct its internal direction signal. Since ~\cite{fu2025jailbreak} additionally report HarmBench-based ASR for GCG, AutoDAN, and SM on the full AdvBench-520 set, we directly use these published results as reference baselines.

To align the evaluation protocol as closely as possible, we use the same AdvBench harmful-behavior set and deterministic/greedy decoding. The results of GCG, AutoDAN, and SM are taken directly from the corresponding AdvBench experiment reported by ~\cite{fu2025jailbreak}, while CFA$^2$ is independently evaluated on the same target models. Since these baselines are reported results rather than reproductions under our implementation, we explicitly mark them as such and treat the comparison as a cross-benchmark reference rather than a strictly controlled reimplementation. In particular, ~\cite{fu2025jailbreak} report the second target model as Mistral-7B-Instruct without specifying the checkpoint revision, so the corresponding Mistral comparison should not be interpreted as strictly checkpoint-matched.

As shown in Table~\ref{tab:advbench}, CFA$^2$ maintains a high attack success rate after AdvBench-specific recalibration and remains competitive with strong existing attacks reported on the same benchmark. These results indicate that the effectiveness of CFA$^2$ is not limited to the specific query distribution of HarmBench or to an intervention direction identified only once; instead, the method can re-establish an effective intervention on a second harmful-query benchmark. Together with the HarmBench results in the main text, this provides additional evidence for the cross-benchmark robustness of CFA$^2$.

\begin{table}[t]
    \centering
    \caption{
    Cross-benchmark attack evaluation on AdvBench using ASR (\%). Results marked with $\dagger$ are reported by ~\cite{fu2025jailbreak} on the full AdvBench-520 harmful-behavior set using the HarmBench classifier. CFA$^2$ is independently recalibrated on AdvBench rather than reusing the intervention direction identified on HarmBench.
    }
    \label{tab:advbench}
    \small
    \setlength{\tabcolsep}{7pt}
    \begin{tabular}{lcc}
        \toprule
        \textbf{Method}
        & \textbf{Llama-2-7B}
        & \textbf{Mistral-7B} \\
        \midrule

        GCG$^\dagger$
        & 47.8
        & 96.9 \\

        AutoDAN$^\dagger$
        & 13.0
        & \textbf{99.5} \\

        SM$^\dagger$
        & 91.7
        & 99.0 \\

        \midrule

        \textbf{CFA$^2$}
        & \textbf{98.5}
        & 99.2  \\

        \bottomrule
    \end{tabular}
\end{table}

\subsection{Transferability Details}
\label{app:transfer_details}
For each harmful request $x$, we construct a shared candidate pool $\mathcal{C}(x)=\{x_1,\dots,x_K\}$ for selection-based methods. The pool contains $K=20$ natural-language variants generated from predefined rewrite families, including role-playing templates, scenario-reframing templates, instruction paraphrases, and semantic rewrites. Random Natural Rewrite, Source-ASR Selection, and CFA$^2$-Selected all select from this same pool. GCG Transfer and AutoDAN Transfer are evaluated separately by directly transferring prompts produced by their original source-model attack procedures.

CFA$^2$ scores each candidate using source-model SAE features. Let $z(x)=\mathrm{SAEEnc}(h_\ell(x))$ denote the SAE activation of prompt $x$ at layer $\ell$. Let $\mathcal{I}_s$ and $\mathcal{I}_u$ denote the task-relevant and defense-related feature sets identified by contrastive analysis, respectively. We define $s(x)=z_{\mathcal{I}_s}(x)$ as the task-feature vector and $\mathrm{Act}_u(x)=\|z_{\mathcal{I}_u}(x)\|_1$ as the defense-feature activation. For each candidate $x_i$, we compute
\[
\mathrm{Score}(x_i)
=
\mathrm{cos}(s(x_i),s(x))
-
\lambda \cdot \mathrm{Act}_u(x_i).
\]
Candidates are ranked by this score. CFA$^2$-Selected transfers the top-ranked candidate for each request. Random Natural Rewrite samples one candidate uniformly from $\mathcal{C}(x)$, while Source-ASR Selection selects the candidate with the highest attack success confidence on the white-box source model.

\subsection{Ablation}
\label{app:ablation}

To better understand the sources of CFA$^2$'s effectiveness, we conduct a series of controlled ablation studies over its main design components. We examine whether sparse representations improve intervention precision, whether the front-door-motivated combination of task-matched contrast and SAE decomposition contributes beyond standard refusal-direction identification, and how sensitive CFA$^2$ is to feature-selection hyperparameters.

\paragraph{Sparse Representations}This section investigates the necessity of sparse representations for precise causal intervention. 
Specifically, we benchmark our proposed SAE-based intervention strategy against a Raw Neuron Baseline, which applies the identical differential analysis and intervention logic directly to the dense activation space without SAE projection.
As illustrated in Table~\ref{tab:ablation_sparsity}, while the raw neuron baseline achieves a non-trivial ASR, it suffers from a drastic degradation in PPL, severely compromising the fluency of the generated text. Meanwhile, the random intervention exhibits negligible efficacy.
This phenomenon empirically corroborates the inherent \textit{polysemanticity} of LLM representations, where intervention may affect task-irrelevant features. 
Consequently, the results demonstrate that {CFA}$^2$ successfully isolates and interrupts the specific causal circuits responsible for the refusal mechanism via sparse disentanglement.
\begin{table}[t]
\centering
\caption{\textbf{Ablation on Sparse Representation.} We compare our SAE-based intervention against a raw neuron baseline (dense space) and a random intervention. While intervening on raw neurons maintains a decent ASR, it causes a catastrophic increase in PPL on output, confirming that polysemanticity in dense representations hinders precise manipulation.}
\label{tab:ablation_sparsity}
\small
{
\begin{tabular}{lcccc}
\toprule
\textbf{Method} & \textbf{Feature Space} & \textbf{ASR (\%)} $\uparrow$ & \textbf{PPL} $\downarrow$ \\
\midrule
Random Intervention & Latent (Sparse) & 5.20 & \textbf{8.55}\\
Raw Neuron Baseline & Dense (Polysemantic) & 48.50 & 652.40\\
\midrule
\textbf{{CFA}$^2$ (Ours)} & \textbf{Latent (Sparse)} & \textbf{98.50} & 9.35\\
\bottomrule
\end{tabular}
}
\vspace{-10pt}
\end{table}




\paragraph{Front-Door-Motivated Representation Decomposition.}
The front-door formulation of CFA$^2$ requires the intervention direction to be identified while preserving task semantics as much as possible and separating representation components that vary with the model's safety behavior. To this end, CFA$^2$ uses task-matched refusal/compliance pairs to constrain task-semantic variation, and further applies SAE decomposition to the mixed representations. Since CFA$^2$ and ORTHO both ultimately use weight orthogonalization for parameter-level intervention, we keep the downstream orthogonalization operator and generation protocol fixed and examine only how different representation-identification strategies affect the intervention direction.

We construct controlled variants along two dimensions: contrast design and representation space. ORTHO uses the original harmful--harmless contrast and identifies the direction in the dense representation space. Matched-ORTHO replaces this contrast with a task-matched refusal--compliance contrast constructed from the same underlying harmful objective, while keeping the dense representation unchanged. SAE-ORTHO retains the original harmful--harmless contrast but identifies the direction in the SAE latent space. The full CFA$^2$ combines task-matched contrast with SAE decomposition. This design allows the contributions of task-matched contrast and sparse decomposition to be examined separately under the same orthogonalization intervention.

As shown in Table~\ref{tab:direction_ablation}, introducing either task-matched contrast or SAE decomposition improves the intervention effectiveness while keeping the downstream intervention unchanged, and combining the two yields the strongest performance. These results indicate that the gain of CFA$^2$ does not arise simply from weight orthogonalization, but is closely tied to the representation decomposition motivated by the front-door formulation: preserving task-related information as much as possible while identifying and attenuating representation components that vary with the safety state. In this sense, contrastive pairing and SAE decomposition respectively provide task-semantic control and separation of safety-related variation, together forming the practical realization of the front-door perspective in CFA$^2$. Orthogonalization then serves as the shared downstream intervention operator that translates the identified direction into a parameter-level modification.

\begin{figure}[t]
    \centering
    \includegraphics[width=0.82\columnwidth]{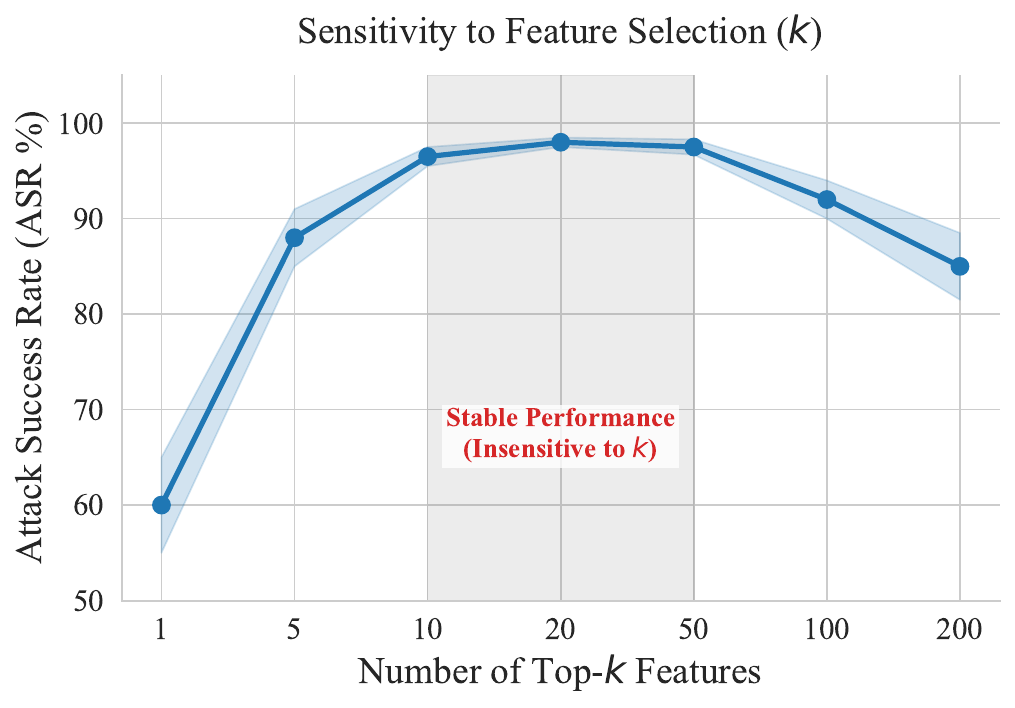}
    \caption{\textbf{Sensitivity to Feature Selection.}
    Attack success rate of CFA$^2$ under different numbers of Top-$k$ SAE features.
    Performance remains stable across a broad range of $k$, particularly for
    $k \in [10, 50]$.}
    \label{fig:ablation_sensitivity}
    \vspace{-10pt}
\end{figure}

\begin{table*}[t]
    \centering
    \caption{
    Ablation of front-door-motivated representation decomposition.
    All variants share the same weight orthogonalization and generation protocol;
    only the intervention-direction identification procedure is varied.
    }
    \label{tab:direction_ablation}
    \resizebox{\textwidth}{!}{
    \begin{tabular}{lccccccc}
        \toprule
        \textbf{Variant}
        & \textbf{Task-Matched Contrast}
        & \textbf{SAE}
        & \textbf{Llama-3.1}
        & \textbf{Llama-2}
        & \textbf{Mistral}
        & \textbf{Gemma-2}
        & \textbf{Avg. ASR $\uparrow$} \\
        \midrule

        ORTHO
        & \xmark
        & \xmark
        & $17.12 \pm 1.58$
        & $22.60 \pm 0.64$
        & $60.30 \pm 1.08$
        & $52.65 \pm 0.96$
        & $38.17 \pm 1.07$ \\

        Matched-ORTHO
        & \cmark
        & \xmark
        & $28.60 \pm 1.21$
        & $34.90 \pm 1.08$
        & $69.20 \pm 0.94$
        & $61.30 \pm 1.16$
        & $48.50 \pm 0.82$ \\

        SAE-ORTHO
        & \xmark
        & \cmark
        & $45.20 \pm 0.96$
        & $49.80 \pm 0.88$
        & $77.60 \pm 0.73$
        & $69.40 \pm 0.91$
        & $60.50 \pm 0.67$ \\

        \textbf{CFA$^2$}
        & \cmark
        & \cmark
        & \textbf{99.40 $\pm$ 0.11} & \textbf{99.10 $\pm$ 0.50} & \textbf{99.70 $\pm$ 0.10} & \textbf{99.60 $\pm$ 0.23} & \textbf{99.45 $\pm$ 0.16} \\

        \bottomrule
    \end{tabular}
    }
\end{table*}
\paragraph{Sensitivity to Hyperparameter $k$}To verify the robustness of CFA$^2$, we investigate the sensitivity of attack efficacy to the candidate pool size $k$(details in Appendix~\ref{app:feature_selection}), aiming to evaluate the recall efficiency of our ranking metric and the noise tolerance of our secondary filtering.
As illustrated in Figure~\ref{fig:ablation_sensitivity}, the ASR rises sharply to saturate at $k \approx 10$ and maintains a wide stability plateau across $k \in [10, 50]$, with only minor degradation observed at extremes due to insufficient recall (small $k$) or excessive noise injection (large $k$).
These results indicate that CFA$^2$ is relatively insensitive to the choice of $k$ within a broad range, with $k=20$ providing a favorable trade-off between feature coverage and intervention stability.




\section{Broader Impact}
\label{app:Broader Impact}
This paper investigates vulnerabilities in large language models (LLMs) to advance safety alignment practices. By exposing these mechanistic weaknesses in widely deployed models, we aim to accelerate the development of robust protective measures before such vulnerabilities can be exploited at scale. While we acknowledge the potential dual-use risks of releasing attack methodologies, we believe that transparent, mechanistic research is indispensable for constructing rigorous safeguards and understanding the inner workings of black-box models.

\end{document}